\documentclass[10pt,twocolumn,letterpaper]{article}

\usepackage{iccv}
\usepackage{times}
\usepackage{epsfig}
\usepackage{graphicx}
\usepackage{amsmath}
\usepackage{amssymb}
\usepackage{soul,color}
\usepackage[T1]{fontenc}
\usepackage{blindtext,nameref}
\usepackage{overpic}
\usepackage{booktabs}
\usepackage{multirow}
\usepackage{float}
\usepackage{stfloats}
\usepackage{marvosym}

\newcommand{\Fref}[1]{Figure~\ref{#1}}

\definecolor{mygreen}{RGB}{20, 190, 10}
\definecolor{myorange}{RGB}{250, 86, 36}
\definecolor{myblue}{RGB}{100, 100, 255}

\usepackage[breaklinks=true,bookmarks=false]{hyperref}

\iccvfinalcopy

\def\iccvPaperID{6600}

\ificcvfinal\fi

\begin{document}

\title{Learning with Noisy Labels for Robust Point Cloud Segmentation}

\author{Shuquan Ye\textsuperscript{1} \quad Dongdong Chen\textsuperscript{2} \quad Songfang Han\textsuperscript{3} \quad Jing Liao\textsuperscript{1}\thanks{Jing Liao is the corresponding author.}\\
\textsuperscript{1} City University of Hong Kong
\quad \textsuperscript{2} Microsoft Cloud AI \quad
\textsuperscript{3} University of California San Diego\\
{\tt\small shuquanye2-c@my.cityu.edu.hk, cddlyf@gmail.com, s5han@eng.ucsd.edu, jingliao@cityu.edu.hk}
}

\maketitle
\ificcvfinal\thispagestyle{empty}\fi

\begin{abstract}
Point cloud segmentation is a fundamental task in 3D. Despite recent progress on point cloud segmentation with the power of deep networks, current deep learning methods based on the clean label assumptions may fail with noisy labels. Yet, object class labels are often mislabeled in real-world point cloud datasets.  In this work, we take the lead in solving this issue by proposing a novel Point Noise-Adaptive Learning (PNAL) framework. Compared to existing noise-robust methods on image tasks, our PNAL is noise-rate blind, to cope with the spatially variant noise rate problem
specific to point clouds
. Specifically, we propose a novel point-wise confidence selection to obtain reliable labels based on the historical predictions of each point. A novel cluster-wise label correction is proposed with a voting strategy to generate the best possible label taking the neighbor point correlations into consideration. We conduct extensive experiments to demonstrate the effectiveness of PNAL on both synthetic and real-world noisy datasets. 
In particular, even with $60\%$ symmetric noisy labels, our proposed method produces much better results than its baseline counterpart without PNAL and is comparable to the ideal upper bound trained on a completely clean dataset. 
Moreover, we fully re-labeled the validation set of a popular but noisy real-world scene dataset ScanNetV2 to make it clean, for rigorous experiment and future research. Our code and data are available at \url{https://shuquanye.com/PNAL_website/}.

\end{abstract}

\section{Introduction}
\label{sec:introduction}

In recent years, the development of deep neural networks (DNNs) has led to great success in 3D point cloud segmentation \cite{3DSemanticSegmentationWithSubmanifoldSparseConvNet,zhao2020point,Ye2021MetaPUAA}.
Thanks to the powerful learning capacity, once high-quality annotations are given, DNNs-based point segmentation methods can achieve remarkable performance. However, such high learning capacity is a double-edged sword, \ie, it can also over-fit label noise and incur performance degradation if there are incorrect annotations.

\begin{figure*}[ht] 
\centering
\setlength{\tabcolsep}{0.5mm}{
\begin{tabular}{cccc}
\includegraphics[width=0.285\textwidth,page=1]{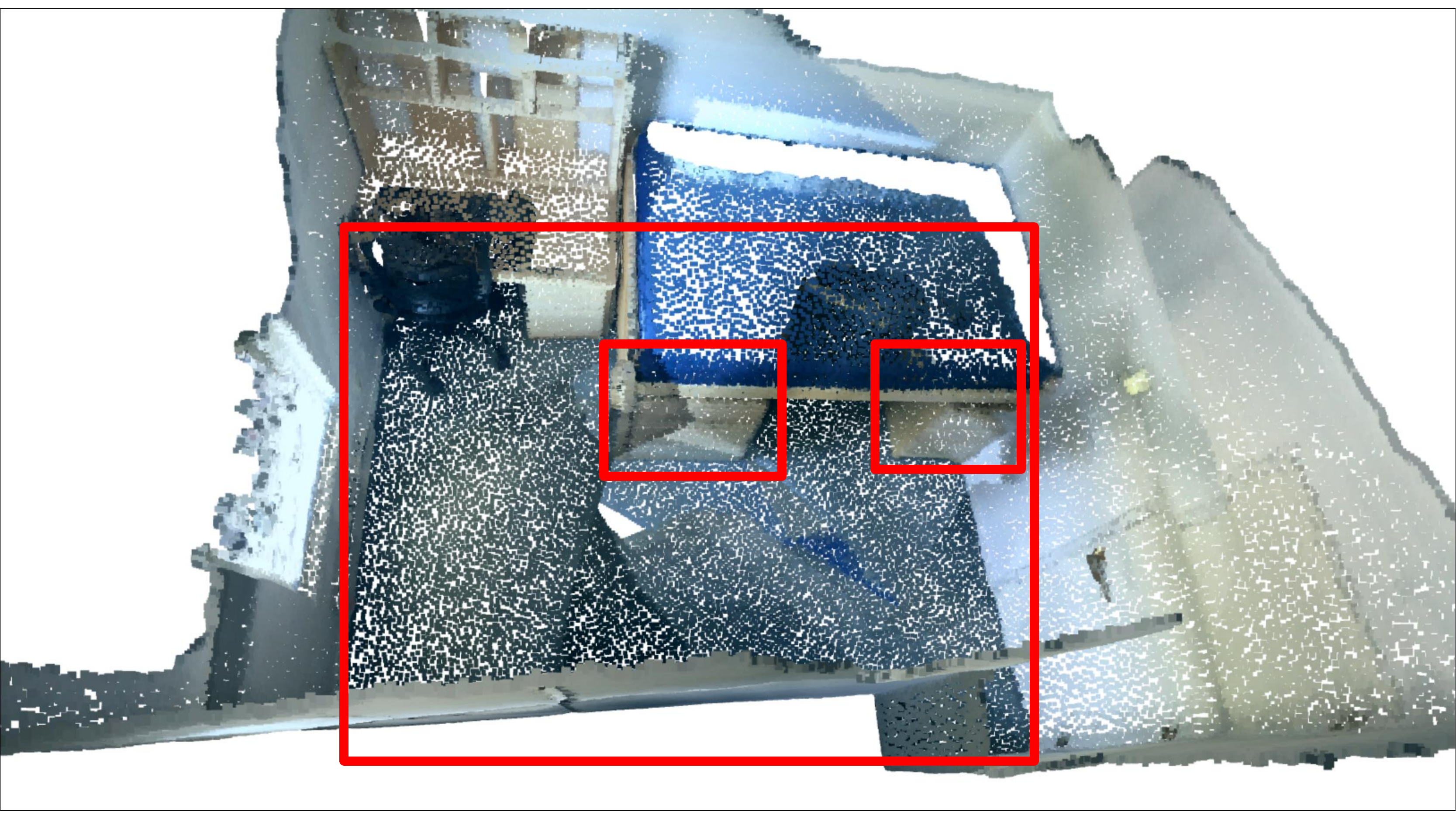} &
\includegraphics[width=0.285\textwidth,page=2]{img/teaser/train_noshuffle_102_oa0.7649055123329163.pdf} &
\includegraphics[width=0.285\textwidth,page=3]{img/teaser/train_noshuffle_102_oa0.7649055123329163.pdf} & \multirow[c]{ 2}{*}[2.3cm]{\includegraphics[width=0.06\textwidth]{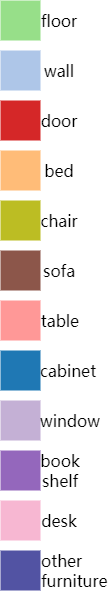}} \\ 
\includegraphics[width=0.285\textwidth,page=1]{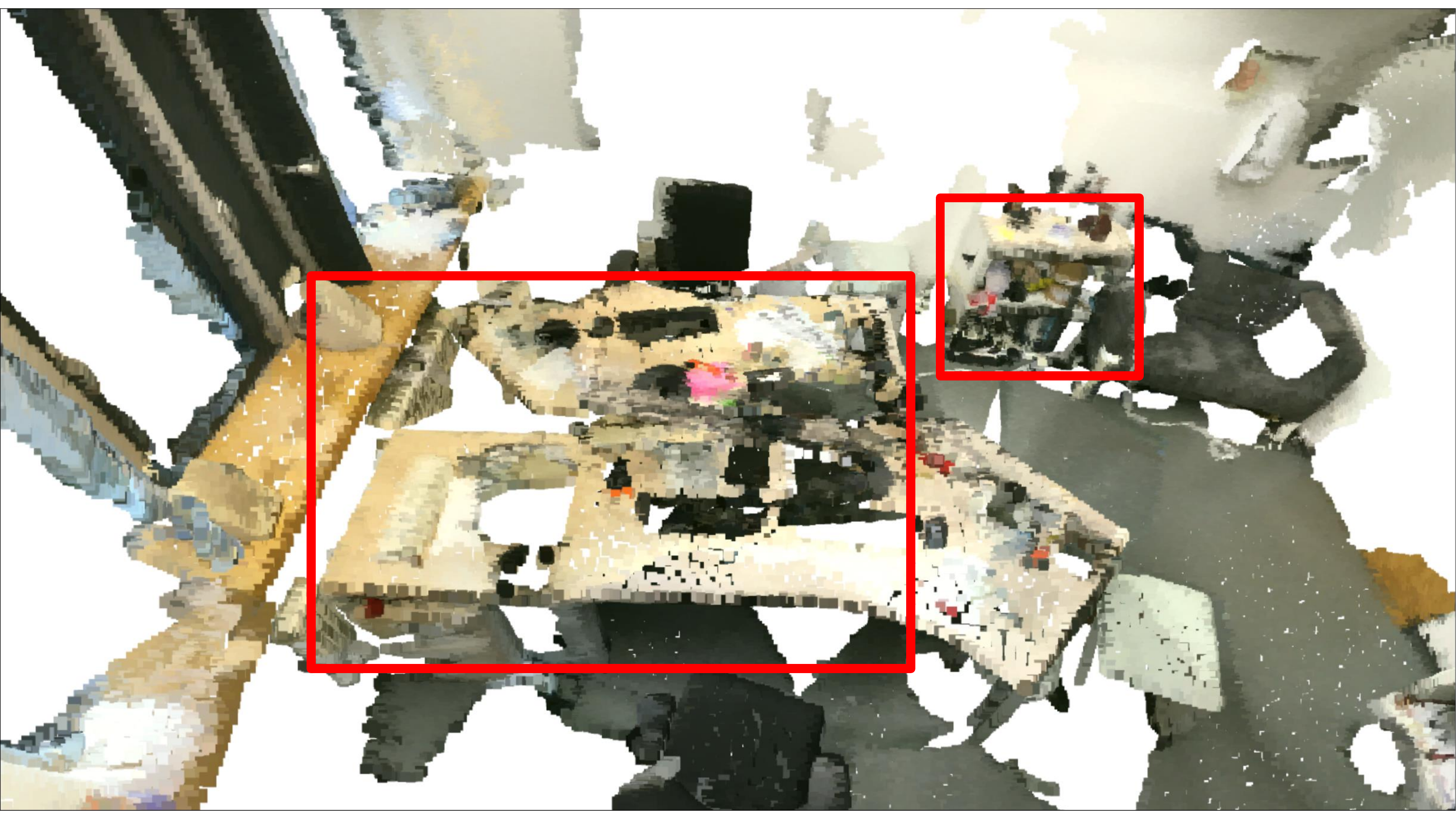} &
\includegraphics[width=0.285\textwidth,page=2]{img/teaser/train_noshuffle_1026_oa0.8556969165802002.pdf} &
\includegraphics[width=0.285\textwidth,page=3]{img/teaser/train_noshuffle_1026_oa0.8556969165802002.pdf} & \\
 Input Scenes & Real-World Noisy GT Labels & Predictions of PNAL & \\

\end{tabular}
}

\caption{Illustration of the instance-level label noise concept in point cloud segmentation. From left to right are the input (noisy instances highlighted red boxes), the manual annotation given by the real-world dataset ScanNetV2, and the prediction of PNAL (more in line with the real category). It is noticeable that this popular dataset suffers from label noise, such as mislabeling the floor as a chair, even that it is already a re-labeled version of ScanNet. Our PNAL framework is trained on this noisy dataset but still achieves correct predictions.}
\label{teaser}
\end{figure*}

In fact, compared to annotating 2D images, clean 3D data labels are more difficult to obtain. It is mainly because: 1) the point number to annotate is often very massive, \eg, million scale in annotating a typical indoor scene in ScanNetV2~\cite{dai2017scannet}; 2) the annotating process is inherently more complex and requires more expertise for the annotators, e.g., constantly changing the view, position and scale to understand the underlying 3D structure. As a result, even the commonly used 3D scene dataset ScanNetV2~\cite{dai2017scannet}, which is already a version after refining the label from the ScanNet, has a large portion of label noise, as shown in \Fref{teaser}.

Based on the above considerations, there is an urgent need to study how to learn with noisy labels for robust point cloud segmentation. However, to the best of our knowledge, most research works about learning with noisy labels focus on image classification, and no previous study exists for point cloud segmentation. More importantly, such works designed for image recognition cannot be directly applied to point cloud segmentation. For example, among the most popular methods, sample selection methods \cite{10.5555/3327757.3327944,yu2019does,shen2019learning,song2019selfie,liu2020early} often assume that the noise rate of all samples is a known constant value. However, the noise rates are often unknown and variable. Robust loss function methods \cite{zhang2018generalized,Wang_2019_ICCV} cannot achieve consistent noise robustness 
to large noise rates. Whereas label correction methods \cite{Reed2015TrainingDN,song2019selfie,ICML2019_UnsupervisedLabelNoise} are designed to correct for image-level label noise, the point cloud segmentation task requires to correct point-level noises. Considering that the point labels within each instance are strongly correlated, directly applying these methods to each point independently without considering the local correlation is suboptimal.

In this paper, we present a novel point noise-adaptive learning (PNAL) framework, which is the first attempt to empower the point cloud segmentation model with resistance to annotation noise.
Specifically, to cope with unknown, possibly heavy, and varying noise rates, we designed a \textbf{point-level} confidence selection mechanism, which obtains reliable
labels based on the historical predictions of each point without requiring a known noise rate. Next, in order to fully utilize the local correlation among labels, we propose a label correction process performed at \textbf{cluster level}. This is done by the proposed \textbf{voting strategy} that tries to merge reliable labels from relevant points to provide the best possible label for each point cluster, with a \textbf{computationally efficient} implementation.

To demonstrate the effectiveness of our PNAL, we compare the proposed framework with various possible baselines based on different network backbones on the synthetic noisy label dataset from stanford large-scale 3d indoor spaces (S3DIS) \cite{armeni_cvpr16}, which shows the great advantage of PNAL on both performance and efficiency.
On the real-world noisy dataset ScannetV2 \cite{dai2017scannet}, we notice that both its training and validation set suffers from the noisy label issue. Therefore, we not only conducted experiments on the original training and validation set. Also, for a more rigorous evaluation, we refine the validation set by manually correcting the noisy labels and evaluate PNAL on this clean set. These results indicate that PNAL is also robust to real-world noise. To further explore PNAL, a complete ablation study, training process analysis, and robustness test were also performed.

To summarize, our contributions are fourfold. 
\begin{itemize}
\setlength{\itemsep}{0pt}
\setlength{\parsep}{0pt}
\setlength{\parskip}{0pt}
\item To the best of our knowledge, this is the first work investigating noisy labels on point cloud data, which has a wide and urgent need for 3D applications where the volume of data is growing drastically.
\item A novel noise-rate blind PNAL framework is proposed to handle spatially variant noise rates in point cloud. It consists of point-level confidence selection, cluster-level label correction with voting mechanism, and can be easily applied to different network architectures.

\item Extensive experiments are conducted to show the clear improvements by PNAL, on both synthetic and real-world noisy label datasets. 
\item We re-labeled the validation set of ScanNetV2 by correcting noises and will make it public to facilitate both point cloud segmentation and noise label learning.

\end{itemize}

\section{Related Work}
\label{sec:related}

\noindent\textbf{Point Cloud Segmentation.}
Given a point cloud, the task of semantic segmentation is to divide it into subsets based on the semantic meaning of the points. Among the related works, point-based networks have recently gained more and more attention. These methods act directly on disordered, irregular, and unstructured point clouds, so that directly applying standard CNNs is not feasible. For this reason, the pioneering work PointNet \cite{qi2017pointnet} was proposed to learn per-point features using cascaded multi-layer perceptrons. It uses cuboidal blocks of fixed arbitrary size to cut the rooms in the scene datasets into blocks when processing datasets. Inspired by PointNet, a series of point-based networks have been recently proposed. In general, following the practice of PointNet, these approaches use octrees \cite{tatarchenko2017octree}, kd-trees \cite{klokov2017escape} or clustering \cite{landrieu2018large} to decompose room scences into room blocks. Those processing method for point cloud data is widely used today \cite{10.1007/978-3-030-58523-5_33}.

\noindent\textbf{Learning with Noisy Labels.}
Different methods have been proposed to train accurate models under noisy labeled data, which can be broadly classified into five categories.

Methods with noise-robust layers \cite{7298885,Goldberger2017TrainingDN,7472164} are intended to model a label transition matrix $\mathrm{T}$. Based on the estimated $\mathrm{T}$, they adjust the output of the network to a more confident label. However, such methods assume a strong correlation between certain labels, which limits the method usages.

Another approach is to design a loss that is robust to noisy labels
\cite{ghosh2017robust,DBLP:conf/iclr/LyuT20}, generalized cross entropy (GCE)\cite{zhang2018generalized}, symmetric cross entropy (SCE)\cite{Wang_2019_ICCV}
, which are popular and easily adaptable to existing architectures. They are originally proposed for classification, but most of them are applicable to segmentation if used pixel-wisely. A recent work \cite{yang2020lncis} proposes to apply the reverse cross entropy~\cite{Wang_2019_ICCV} for the foreground-instance sub-task and a standard cross entropy (CE) loss for the foreground-background in instance segmentation. A major limitation is that they cannot handle heavy noisy labels. 

Loss adjustment methods \cite{Reed2015TrainingDN,NIPS2017_2f37d101,NEURIPS2018_ad554d8c,ICML2019_UnsupervisedLabelNoise,song2019selfie,zhang2020characterizing,wang2020noise} reduce the negative impact of noisy labels by adjusting the loss of all training samples. In general, while these methods fully exploring the training data, they take the risks of false correction. 
Among them, \cite{zhang2020characterizing,wang2020noise} are the latest methods that try to solve the noisy label problem on binary segmentation. 

To avoid false correction, sample selection methods \cite{10.5555/3327757.3327944,yu2019does,shen2019learning,song2019selfie,liu2020early} select true-labeled samples from noisy data. 
However, they take the risk of discarding usable samples, and require either a true noise rate or a fully clean validation set. Our approach is a hybrid of sample selection and loss correction. Unlike other sample selection methods, no noise-rate is required. Different from loss correction, instead of correcting all samples, we correct labels according to confidence considering its locally similar region.

\section{Problem Description}
\label{sec:taskdef}
We define the task of multi-class point cloud semantic segmentation with noisy labels. 
Formally, we denote point cloud data as $\mathbf{X} \in \mathcal{R}^{N \times C}$ of $N$ points with $C$ features of coordinates and RGB values possibly, and its semantic label as $\mathbf{Y}$, and $M$ as the class number. Our target is to train a model $\mathbf{f}_{\theta}(\mathbf{X})$ robust to the label noise in the training set.

We observe that while the semantic label of an instance might be wrong, we seldom encounter incorrect instance partitioning. For example, a table may be labeled as a sofa, but we seldom see a table that is partially labeled as a sofa. This implies that our method should correct the noise at the instance level.  However, the ground truth instance information may not available, and predicting object instances itself is a challenging task.

Alternatively, we create a cluster based noise correction method, where each cluster consists of a local patch of points. The main assumption is that points in a cluster are considered to belong to the same instance. 

This cluster can be generated with off-the-shelf clustering algorithms. In experiments, we mainly use the DBSCAN~\cite{ester1996density}. 
DBSCAN is a density-based clustering algorithm that does not require the specification of the cluster number in the data, unlike k-means. DBSCAN can find arbitrarily shaped clusters, and this characteristic makes DBSCAN very suitable for LiDAR point cloud data. 
In the experiment we also tried another clustering method GMM.

During training, we correct the label on a cluster level. Based on the experiments described in Sec.~\ref{subsec:abl2_vsEpsandGMM}, we demonstrate that our method is robust to the granularity of cluster to a certain range and insensitive to clustering methods. We denote the point cloud and its label in a cluster $\mathbf{C}_{i}$ as $(X_{\mathbf{C}_{i}}, Y_{\mathbf{C}_{i}}), 1 \leq i \leq k$, and $k$ is the number of clusters.

\section{Methodology}
\label{sec:methodology}

\subsection{Pipeline Overview}

\begin{figure*}[ht]
\includegraphics[width=\textwidth]{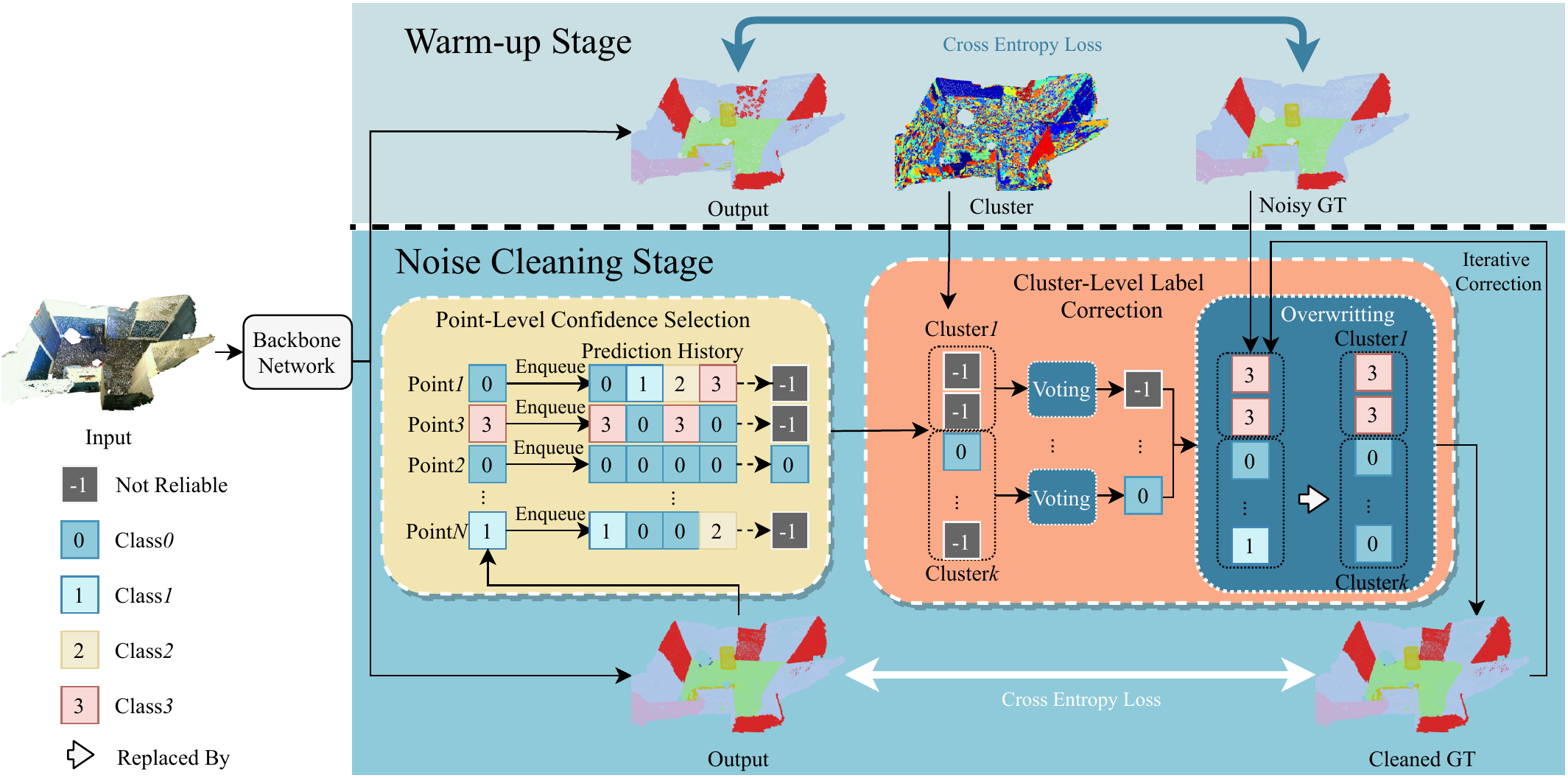}
\caption{System pipeline. In the warm-up stage, the network is updated with CE as usual.
In the noise cleaning stage, we enqueue the output to the prediction history and point wisely perform confidence selection to get reliable labels. With these results, we do voting at cluster level, then correct the original noisy GT or the previously cleaned GT. Finally the obtained cleaned GT guides the network update.
}
\label{Fig:pip}
\end{figure*}

The overall framework of our proposed method is illustrated in Fig.~\ref{Fig:pip}. 
Our training procedure is divided into two stages, a warm-up stage introduced in Sec.~\ref{sec:warmup} and a noise-cleaning stage introduced in Sec.~\ref{sec:selection} and Sec.~\ref{sec:correction}. We first train the network on all samples by default manner for $E_{warm-up}$ epochs, where $E_{warm-up}$ denotes the number of epochs for this stage. The warm-up stage motivates the network to learn the easy data, which are largely consistently labeled correct data.
 
Then we further train the network with our proposed noise-cleaning stage. The main idea is to identify the potential noisy label based on the behavior of the network prediction and update the selected data label to a more reliable label.  From the observation that the ground truth label is often corrupted on an instance-level, we encourage to correct data labels in a group-wise manner. That we propose to first cluster point cloud into small patches and then update point label patch-by-patch. Point cloud in one patch is set to the same semantic label. Besides, we predict the new label of each cluster through a voting strategy.

\subsection{Warm-Up}\label{sec:warmup}
According to the study of memorization effects \cite{arpit2017closer}, deep neural networks are prone to learn clean, easy samples first, and then other noisy samples gradually, even in the presence of noisy labels. Therefore, in the warm-up stage, we adapt no strategies and train the network with a common cross-entropy loss. The detailed formula of $\operatorname{Loss}_{warm-up}$ is:

\begin{equation} \label{eq:1}
\operatorname{Loss}_{warm-up}=-\frac{1}{B} \sum_{i=1}^{B} \sum_{m=0}^{M} q\left(m \mid \mathbf{X}_{i}\right) \log p\left(m \mid \mathbf{X}_{i}\right),
\end{equation}
where $B$ is the number of samples inside a mini-batch, $ p\left(m \mid \mathbf{X}_{i}\right)$ denotes classification confidence of each class $m \in \{1,...,M\}$, and $q\left(m \mid \mathbf{X}_{i}\right) \in {\{0,1\}}^{N \times M}$ is one-hot encoded label. 

The warm-up stage power the network with easily learned data samples. However, if trained with a large number of epochs for the warm-up stage, the network will tend to fit noisy labels. Let $E_{clean}$ denotes the number of epochs for the noise-cleaning stage. In noise-cleaning stage of training, remarkably, no hyperparameters such as noise rate are required. 
We observe that, the precision of both the replaced labels and the correctly fixed labels increases gradually as the training step progresses, and label replacement eventually expands to almost the entire training set, as analysized in \ref{subsec:corrprocess}.
Moreover, through our analysis in \ref{subsec:abl1_Ewarmupvsnoiserate} on $E_{warm-up}$, too much warm-up causes the model fitting more noisy data, which affects the performance when the noise rate is large. Meanwhile, the optimal $E_{warm-up}$ setting is not sensitive to the noise rate. At different noise rates, we can derive the following relationship between the optimal $E_{warm-up}$ and $E_{clean}$,

\begin{equation} \label{eq:2}
E_{warm-up}=\frac{1}{5}E_{clean}.
\end{equation}

\subsection{Point-Level Confidence Selection}\label{sec:selection}
In this part, we aim to select reliable samples from each mini-batch and obtain the confidential labels for these samples that can be corrected with high probability.

Previous studies of sample selection \cite{song2019selfie,10.5555/3327757.3327944} adopt the widely used loss-based separation which tries to construct clean set $\mathcal{C}$ by selecting $(1-\tau) \times 100\%$ of low-loss instances, where $\tau$ is the noise rate. While significant improvements have been achieved in these works, such an approach faces two key problems. First, it is not appropriate to assume that the noise rate is available and constant on the 3D point cloud. Second, it would exclude $\mathcal{C}_{h}$ which tends to produce high losses, making it difficult for hard cleans to participate in the network update. To address the above weaknesses of the loss-based sample selection methods, instead of selecting clean samples, we propose to directly select the reliable samples without knowing the noise rate.

Inspired by the bootstrapping \cite{Reed2015TrainingDN}, we develop a confidence point selection method for dynamically selecting reliable samples and target labels without requiring explicit noise rate. Instead of using predicted class probabilities to generate regression targets, a new criterion, inspired by SELFIE \cite{song2019selfie}, is designed based on history prediction. In detail, a sample with consistently predicted label is regarded as the reliable sample, and its most frequently predicted label is the reliable label, as defined below:

\textup{Definition 1.} A point $x_{{n}}$ is a reliable sample if the predictive confidence $F(x_{{n}} ; q)$ satisfies $F(x_{{n}} ; q) \geq \sigma (0 \leq \sigma \leq 1)$.
The predictive confidence is defined as 
\begin{equation} \label{eq:2}
F(x ; q)=(1 / \lambda) \text { entropy }(P(m \mid x ; q))
\end{equation}
where $\lambda = - \log (\frac{1}{M})$ is a normalization term for normalizing to $[0,1]$.
We denote the predicted label of the sample point $x$ at time $t$ as
$\hat{m}_{t}=\mathbf{f}_{{\theta}_{t}}(x)$. Then, the label history of the sample $x$ that stores the predicted labels of the previous $q \leq E_{warm-up}$ times is $H_{x}(q)=\left\{\hat{m}_{t_{1}}, \ldots, \hat{m}_{t_{q}}\right\}$, where $q$ is history length. 
Next, as given in Eq. \ref{eq:3}, $P(m \mid x ; q)$ is the probability of the label $m \in\{1,\ldots, M\}$ estimated as the label of the sample point $x$. 
\begin{equation} \label{eq:3}
P(m \mid x ; q)=\frac{\sum_{\hat{m} \in H_{x}(q)}[\hat{m}=m]}{\left|H_{x}(q)\right|}
\end{equation}
, where $[\cdot]$ is the Iverson bracket notation. Then we denote the set of reliable samples as $\mathbf{X}_{reliable}$.
Finally, we define the reliable label as $m^*_{{n}}$ where
$$
m^*_{{n}}=\arg \max_m P(m \mid x_{{n}} ; q).
$$

Specifically, our method differs from SELFIE in three significant ways. 
First, our method is noise-rate blind, to handle spatially variant noise-rate problem in point cloud. 
Second, our confidence selection mechanism is point-level, and our label correction process is done on cluster level with the help of a novel voting strategy, in order to consider local relationship between point labels. Note that the reliable label in selection stage may not be the label won for correction process.
Last but not least, our detailed implementation has to be much more computationally efficient, to allow for point-level selection and cluster-level correction.

\subsection{Cluster-Level Label Correction}\label{sec:correction}

The label correction process is to selectively replace the label with the best possible label in locally similar regions. Ideally, the local similarity region is defined by the ground-truth instance. However, since instance labels may not be available in practice, we use cluster as an alternative. From each cluster $\mathbf{C}_{i} (1 \leq i \leq k)$, as given in Eq. \ref{eq:4}, the ones containing reliable samples will be selected for the label correction in subsequent steps.

\begin{equation} \label{eq:4}
\{ \mathbf{C}_{i} \mid \exists x_{{n}}: x_{{n}} \in \mathbf{X}_{reliable} \land x_{{n}} \in \mathbf{X}_{\mathbf{C}_{i}} \}
\end{equation}
, where $n \in\{1,\ldots, N\}$. 
Next, for each of these clusters, \eg $\mathbf{C}_{i^*}$, we will replace the label $\mathbf{Y}_{\mathbf{C}_{i^*}}$ with a best label locally. The goal label can be found by the proposed voting strategy according to the overall label occurrences within the cluster. We use $occ^{m}_{i^*}$ to denote the occurrence number of reliable samples with $m^*_{{n}}=m$.

\begin{equation}
\begin{aligned}
occ^{m}_{i^*} {} & = \sum_{n=1}^N [m^*_{{n}}=m \land x_{{n}} \in \mathbf{X}_{\mathbf{C}_{i^*}}] \\
        & =\sum_{n=1}^N [m^*_{{n}}=m] [x_{{n}} \in \mathbf{X}_{\mathbf{C}_{i^*}}]
\end{aligned}
\end{equation}
, where $[\cdot]$ is the Iverson bracket.
Then, occurrences for each class are formed as a vector $occs_{i^*} = \left( occ^1_{i^*}, ..., occ^{M}_{i^*} \right)$, and we can find the top occurrence by $occs^{top}_{i^*} = \max_{1\leq m \leq M} occs^{m}_{i^*}$ . A winner label is randomly chosen from $ \{m \mid occ^{m}_{i^*} \geq \frac{ occs^{top}_{i^*} }{\gamma} \}$ to overwrite the label of this cluster.
Note that, in a special case of $\gamma=1$, the winner label in this cluster will be from the top reliable labels $\{m \mid occ^{m}_{i^*} = { occs^{top}_{i^*} } \}$.
According to our ablation study, $\gamma=4$ achieves best performance. Note that the original label may not be mislabeled or different from the winner label. And the labels may be repeatedly overwritten during the training process. Finally, we update the network with these replaced labels by a cross-entropy loss, and the samples whose label have never been replaced are not involved in the gradient calculation.

\section{Experiments}

\subsection{Datasets and Noise Settings}
\label{sec:dataandnoise}
To demonstrate the effectiveness of our proposed method, we conduct experiments on two datasets, ScanNetV2 \cite{dai2017scannet} and S3DIS \cite{armeni_cvpr16}. ScanNetV2 is a popular 3D real-world dataset with label noise. S3DIS is a commonly used scene dataset with much cleaner labels, which can be considered as clean data. There we can artificially build noisy datasets from S3DIS with various noise settings.

\noindent\textbf{ScanNetV2.} The ScanNetV2 3D segmentation dataset contains $1,513$ annotated rooms with $21$ semantic elements in total. According to the scan annotation pipeline for ScanNetV2 \cite{dai2017scannet,smartscenes}, a normal-based graph cut image segmentation method \cite{felzenszwalb2004efficient} is first utilized to get a basic, pre-segmentation. These provide a reliable reference to an object instance, however, the class label can be carelessly mislabeled in practice.
Also, since the rooms are distributed to different annotators, inconsistent labels can be found even for the same object with different placements in the same scene.
These observations match our assumption about the noise pattern in point cloud segmentation that mislabeling occurs at the object-instance-level. 
And we noticed that the noisy label problem also exists even in its validation sets. Since this issue has never been mentioned in other studies, we manually correct such noise labels of all scenes from the validation sets of ScanNetV2 for more accurate evaluation.
Note that we did not perform evaluation on the benchmark test split, due to its unknown noise rate and unavailable annotation.

\noindent\textbf{S3DIS.} S3DIS contains point clouds of $272$ rooms in $6$ large-scale indoor scenes in three buildings with $12$ semantic elements. 
The instance label is borrowed from \cite{pham-jsis3d-cvpr19} which are manually annotated. 
Compared to ScannetV2, the S3DIS dataset has a much smaller amount of scenes, less scene complexity, and fewer classes, and the errors occurring in class labels are clearly less than the former. Therefore, we treat the S3DIS dataset as a completely clean dataset, i.e., the noise rate is $0$.

We generate a noisy dataset from S3DIS by randomly changing the point label at an object-instance level, guided by our noise pattern assumption.
Following previous work on image classification with noisy data, we model the noisy dataset with two noise types: symmetric and asymmetric. For symmetric noise, the point label is modified to other labels with equal probability at instance level. 
Also, we found that some class pairs are easily mislabeled as each other
in the real-world noisy ScanNetV2, such as door and wall, while some pairs are not that confusable, such as wall and desk. Based on this, we create asymmetric noisy S3DIS dataset, mimicking the way in real-world. 
In particular, we identify the easily misclassified label pairs, including door-wall, board-window, sofa-chair, and randomly flip the label inside each label pair with a probability of $\tau_{pair}$. To note that ours setting is different from previous work on image noisy labels. On point cloud data with only $12$ semantic classes, it is inappropriate and unrealistic, to define another confusable class for all classes. Therefore, for classes without pairs, we follow the symmetrical noise setting, to achieve a specified value of the overall noise rate $\tau$. We will show results in the supplemental file following the asymmetric noise setting in previous works on image.

\subsection{Implementation Details}
The DBSCAN algorithm is used for point cloud segmentation in this study, if not specifically stated, with $\varepsilon=0.018$. 
For the real-world noisy dataset ScanNetV2 and artificially created noisy dataset S3DIS, room scenes are divided into room blocks of size $1.0 \times m$ and stride $0.5 \times m$. We random sample $4096$ points for each room block during training. 
We report the results in terms of overall accuracy (OA) and mean intersection over union (mIoU) with DGCNN~\cite{dgcnn}, Pointnet2~\cite{qi2017pointnetplusplus}, and SparseConvNet~\cite{3DSemanticSegmentationWithSubmanifoldSparseConvNet} as backbones.
Without special notation, all experiments are conducted with DGCNN as backbone. 
For symmetric noise, we conduct experiments on noise rates $\tau \in \{20\%,40\%,60\%,80\%\}$. 
For asymmetric noise, we test on a large noise rate $\tau = 60\%, \tau_{pair} = 40\%$. 
All the results on S3DIS are tested on the clean 6th-Area. We train a total of 30 epochs,  including the warm-up stage and clean-noise stage. The history length is set to 4.

\subsection{Baselines}
Note that we are the first handling noisy label on point cloud segmentation. We try our best to adapt previous works and create the following baselines: CE, GCE\cite{zhang2018generalized}, SCE\cite{Wang_2019_ICCV} and SELFIE\cite{song2019selfie}. 
The first three methods can be naturally applied to the point cloud segmentation as point level guidance.
To adopt SELFIE, we apply the original implementation of image-level SELFIE point-wisely
. We experimentally find that the optimal warm-up threshold is $5$. The other settings are the same as in their paper.

\subsection{Performance Comparison on S3DIS}

\begin{table*}[ht]
\centering
\setlength{\tabcolsep}{5.0mm}{
\begin{tabular}{@{}c|c|cccc|c@{}}
\toprule
\multirow{2}{*}{Methods} & \multirow{2}{*}{0\%} & \multicolumn{4}{c|}{Symmetric Noise ($\tau$)}                                  & Asymmetric Noise \\ \cmidrule(l){3-6} 
                         &                      & 20\%            & 40\%            & 60\%            & 80\%            & $\tau_{pair}=40\%$,$\tau=60\%$             \\ \midrule 
DGCNN\cite{dgcnn}+CE                 & \textbf{0.8692}      & 0.7506          & 0.6732          & 0.6390          & 0.5060          & 0.5634           \\
DGCNN\cite{dgcnn}+SCE\cite{Wang_2019_ICCV}                & 0.7768               & 0.7524          & 0.7230          & 0.6509          & 0.5705          & 0.7084           \\
DGCNN\cite{dgcnn}+GCE\cite{zhang2018generalized}               & 0.7067               & 0.7003          & 0.6997          & 0.6967          & 0.6880          & 0.6614           \\
DGCNN\cite{dgcnn}+SELFIE\cite{song2019selfie}*            & 0.8673               & 0.8158          & 0.7914          & 0.7725          & 0.7163          & 0.7500           \\
DGCNN\cite{dgcnn}+PNAL               & 0.8686               & \textbf{0.8569} & \textbf{0.8378} & \textbf{0.8236} & \textbf{0.7651} & \textbf{0.7968}  \\ \midrule
PointNet2\cite{qi2017pointnetplusplus}+CE             & \textbf{0.8898}      & 0.7008          & 0.6796          & 0.5850          & 0.5204          & 0.5648           \\
PointNet2\cite{qi2017pointnetplusplus}+PNAL           & 0.8852               & \textbf{0.8385} & \textbf{0.8271} & \textbf{0.8067} & \textbf{0.7708} & \textbf{0.8202}  \\ \bottomrule
\end{tabular}
}
\caption{OA Comparison of different methods on artificially created noisy S3DIS. The tops with different backbones are shown in bold.}
\vspace{-0.4em}
\label{Tab:s3dis}
\end{table*}

Table.~\ref{Tab:s3dis} shows the performance of baselines and PNAL under different backbones, noisy rates and noise types.  

The first five rows show the results with DGCNN backbone. 
In the case of DGCNN+CE, its performance drops quickly by 11.86\% at 20\% noise rate alone and by 23\% at 60\% symmetric noise rate compared to the result on clean training data.  This shows that label noise hurt the training process severely.
We observe that previous noise-robust methods SCE and GCE perform no better or worse than CE at 0\% and 20\% noise rates, and only 1.19\% and 5.77\% improvement at 60\% symmetric noise rate. 
The noise correction method SELFIE helps improve the performance by 13.35\% at 60\% symmetric noise rate.  
These are expected since these methods works with small or constant noise rates, while point cloud training suffers from extreme noise rate variation. And they do not consider the label correlation of local regions, so it is difficult to achieve optimal results.
Compared to the DGCNN+SELFIE framework, DGCNN+PNAL shows a further improvement of more than 4.11\% for all noise settings.  It is worth noting that SELFIE requires a noise rate and takes more than 10 hours per epoch on average, while DGCNN+PNAL takes only 3 hours and 51 minutes. We owe this to the noise-blind pipeline, and the voting design that takes into account the label correlation of local regions. Furthermore, our method significantly improves result by 10.63\%, 16.46\%, 18.46\%, 25.91\%, 23.34\% at 20\%, 40\%, 60\%, 80\% symmetric noise and asymmetric noise, respectively.  
Consistently, our performance improves by 13.77\%, 14.75\%, 22.17\%, 25.04\%, and 25.54\% with PointNet2 as backbone, as shown in the last two rows.

\begin{figure*}[t] 
\centering
\setlength{\tabcolsep}{0.2mm}{
\begin{tabular}{ccccc}
\includegraphics[width=0.190\textwidth,page=1]{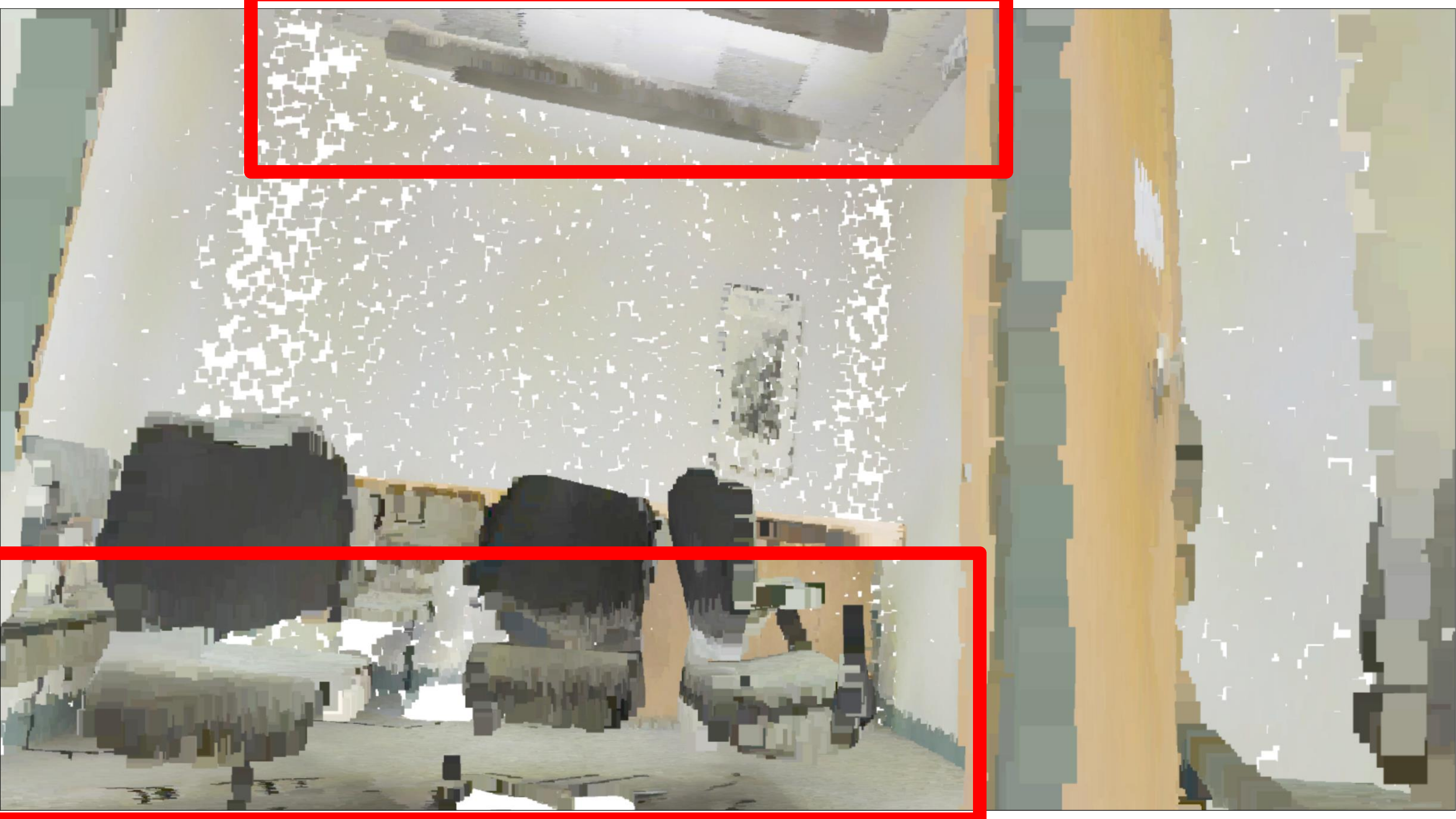} &
\includegraphics[width=0.190\textwidth,page=2]{img/s3dis_our_vs_dgcnn/Area_6_conferenceRoom_1.pdf} &
\includegraphics[width=0.190\textwidth,page=3]{img/s3dis_our_vs_dgcnn/Area_6_conferenceRoom_1.pdf} & 
\includegraphics[width=0.190\textwidth,page=4]{img/s3dis_our_vs_dgcnn/Area_6_conferenceRoom_1.pdf}  &  \multirow[c]{3}{*}[1.7cm]{\vspace{1mm} \includegraphics[width=0.07\textwidth]{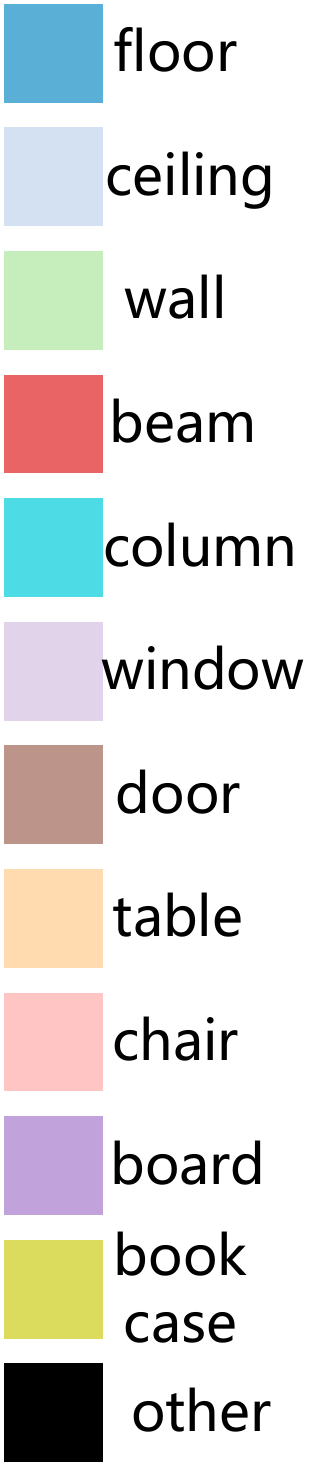}} \\ 
\includegraphics[width=0.190\textwidth,page=1]{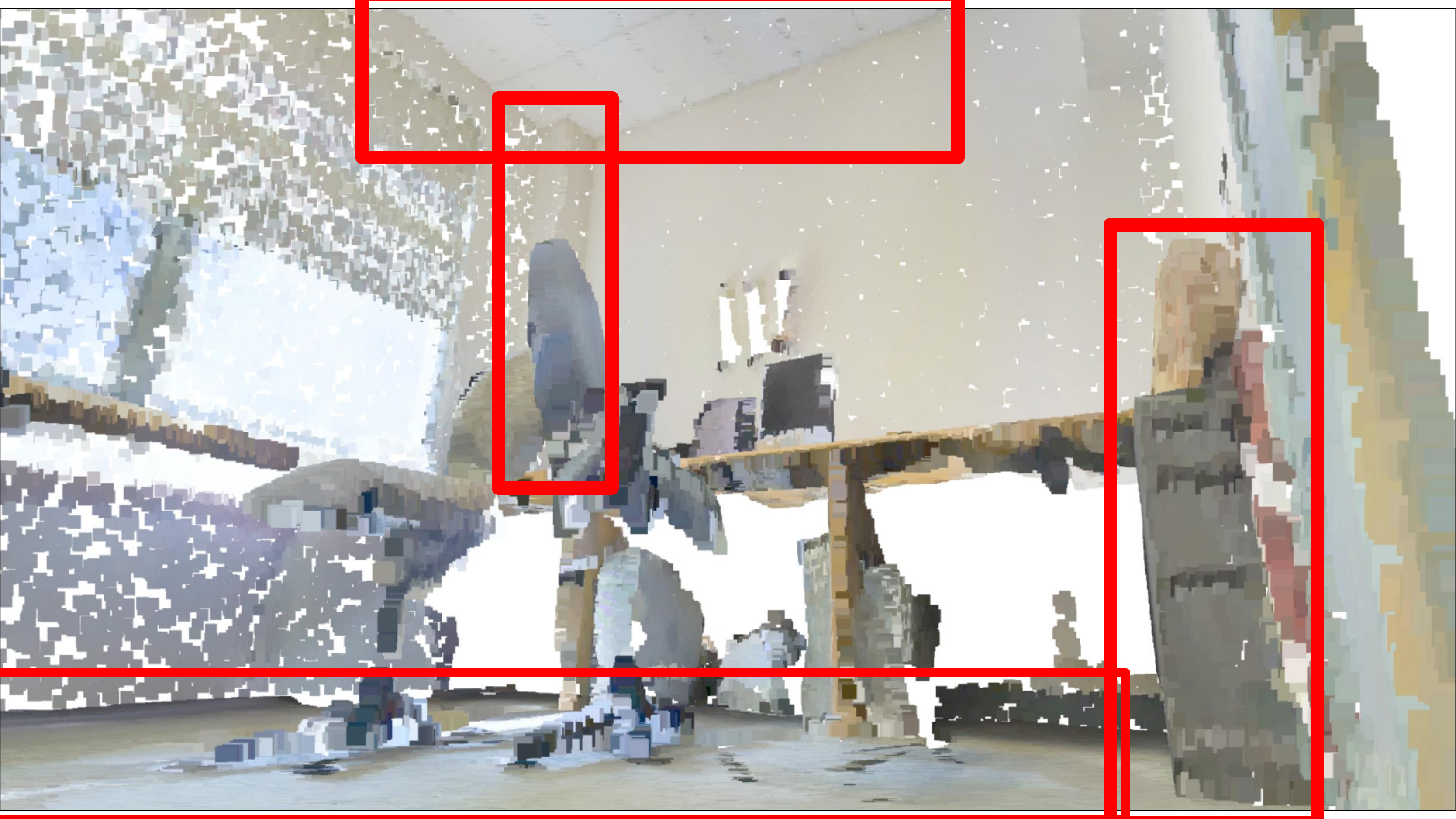} &
\includegraphics[width=0.190\textwidth,page=2]{img/s3dis_our_vs_dgcnn/Area_6_office_30.pdf} &
\includegraphics[width=0.190\textwidth,page=3]{img/s3dis_our_vs_dgcnn/Area_6_office_30.pdf} & 
\includegraphics[width=0.190\textwidth,page=4]{img/s3dis_our_vs_dgcnn/Area_6_office_30.pdf} \vspace{-1mm} & \\
\includegraphics[width=0.190\textwidth,page=1]{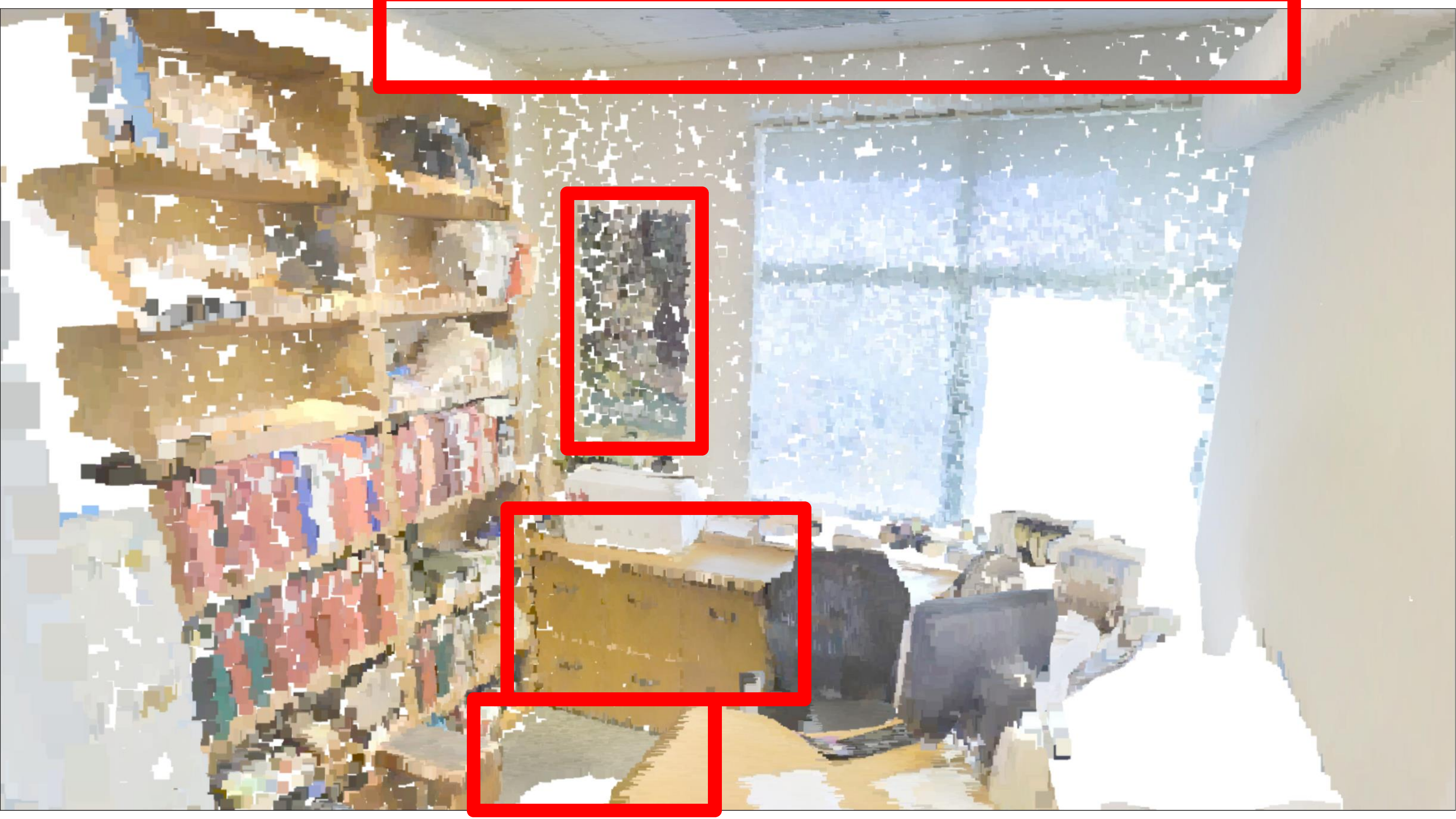} &
\includegraphics[width=0.190\textwidth,page=2]{img/s3dis_our_vs_dgcnn/Area_6_office_31.pdf} &
\includegraphics[width=0.190\textwidth,page=3]{img/s3dis_our_vs_dgcnn/Area_6_office_31.pdf} & 
\includegraphics[width=0.190\textwidth,page=4]{img/s3dis_our_vs_dgcnn/Area_6_office_31.pdf} \vspace{-1mm} & \\
 Input Scenes & GT Labels & DGCNN+CE & DGCNN+PNAL\\

\end{tabular}
}
\caption{From left to right: Scenes in S3DIS testset, clean GTs, predictions of DGCNN+CE, and DGCNN+PNAL.}
\label{Fig:s3distest}
\vspace{-1em}
\end{figure*}

\subsection{Performance on ScanNetV2}

\begin{figure*}[t] 
\centering
\setlength{\tabcolsep}{0.2mm}{
\begin{tabular}{ccccc}
\includegraphics[width=0.190\textwidth,page=1]{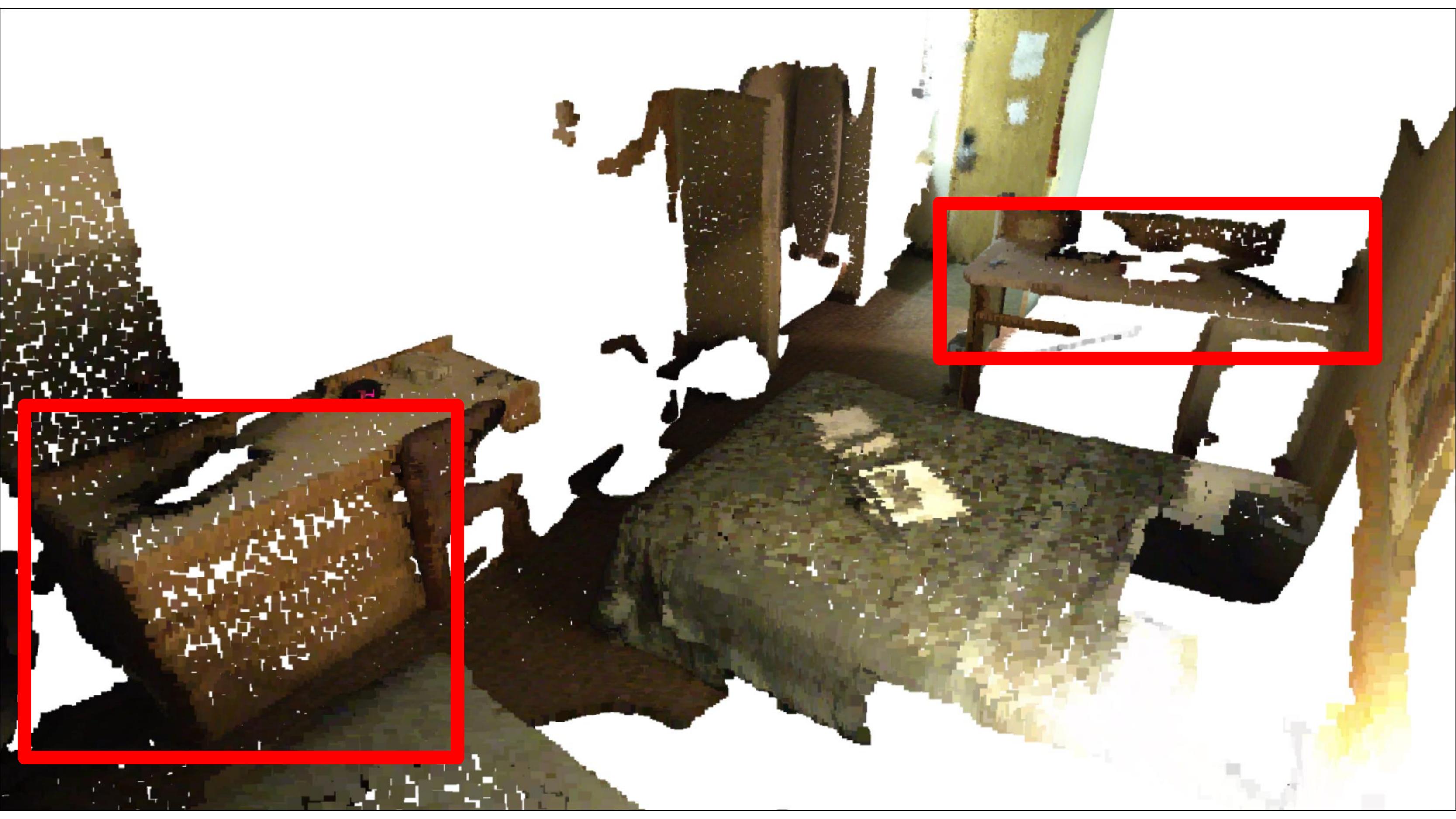} &
\includegraphics[width=0.190\textwidth,page=2]{img/scannetv2_our_vs_gt_vs_scn/72_oa0.8834_oa20.9227.pdf} &
\includegraphics[width=0.190\textwidth,page=3]{img/scannetv2_our_vs_gt_vs_scn/72_oa0.8834_oa20.9227.pdf} & 
\includegraphics[width=0.190\textwidth,page=4]{img/scannetv2_our_vs_gt_vs_scn/72_oa0.8834_oa20.9227.pdf} & \multirow[c]{ 3}{*}[1.7cm]{\includegraphics[width=0.07\textwidth]{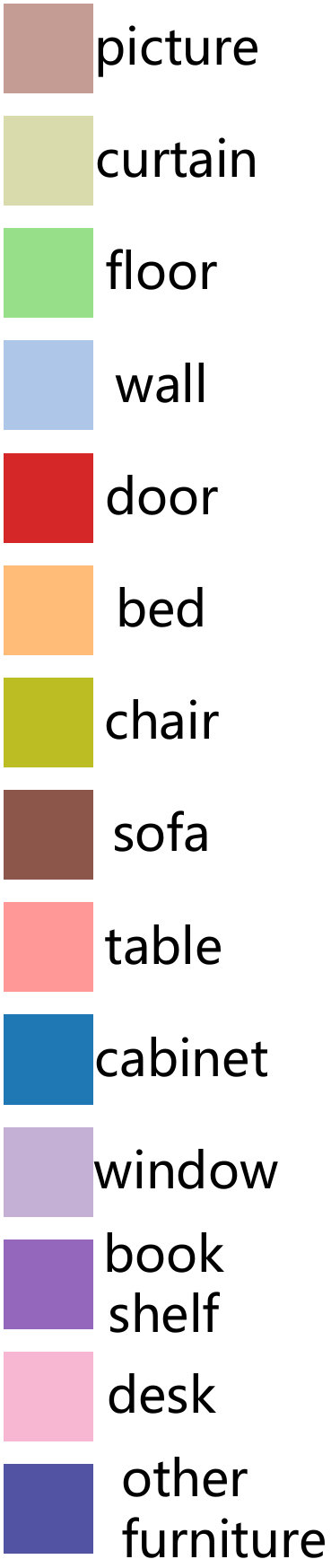}} \vspace{-1mm} \\ 
\includegraphics[width=0.190\textwidth,page=1]{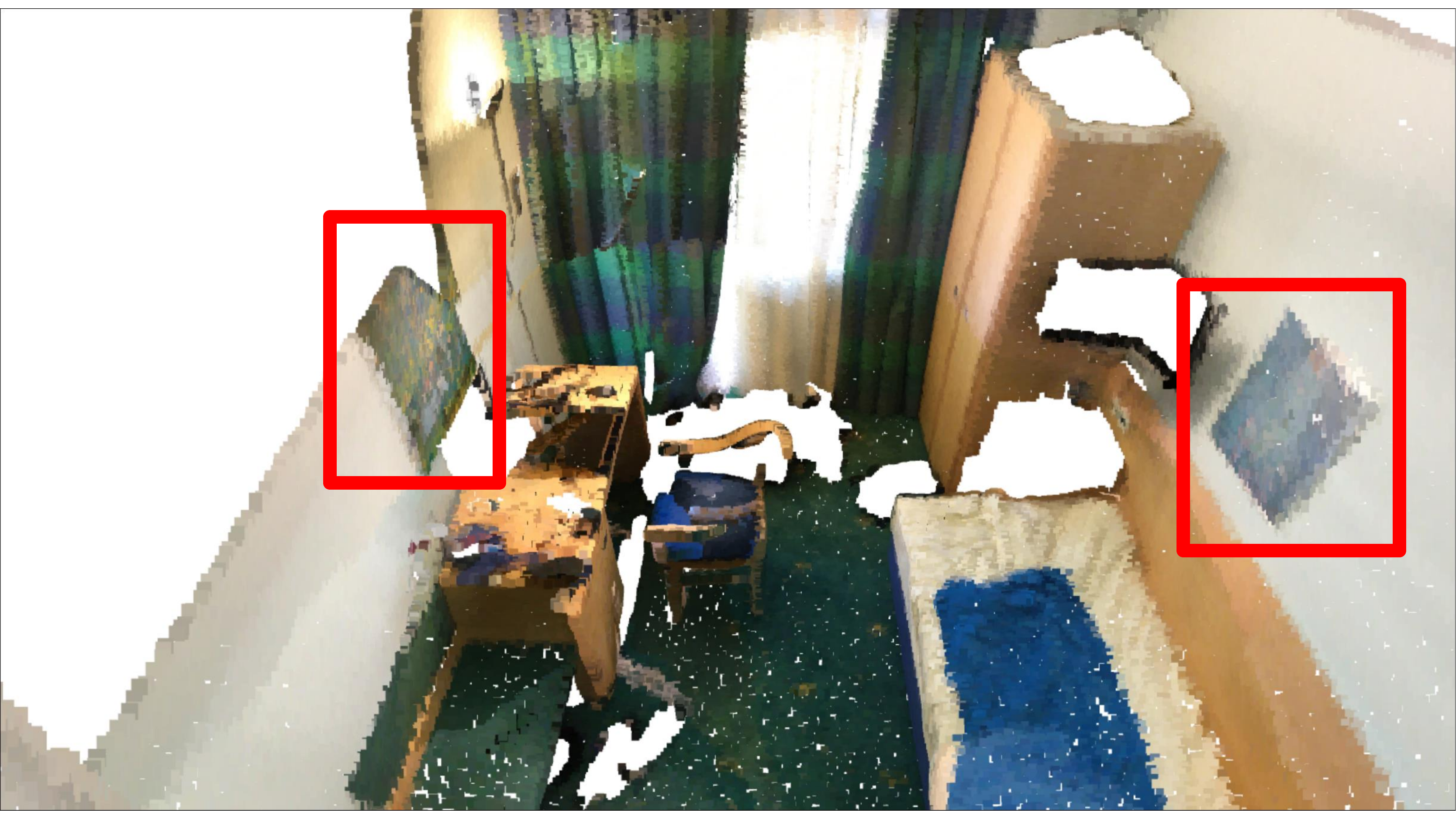} &
\includegraphics[width=0.190\textwidth,page=2]{img/scannetv2_our_vs_gt_vs_scn/13_oa0.8621_oa20.9072.pdf} &
\includegraphics[width=0.190\textwidth,page=3]{img/scannetv2_our_vs_gt_vs_scn/13_oa0.8621_oa20.9072.pdf} & 
\includegraphics[width=0.190\textwidth,page=4]{img/scannetv2_our_vs_gt_vs_scn/13_oa0.8621_oa20.9072.pdf} \vspace{-1mm} & \\
\includegraphics[width=0.190\textwidth,page=1]{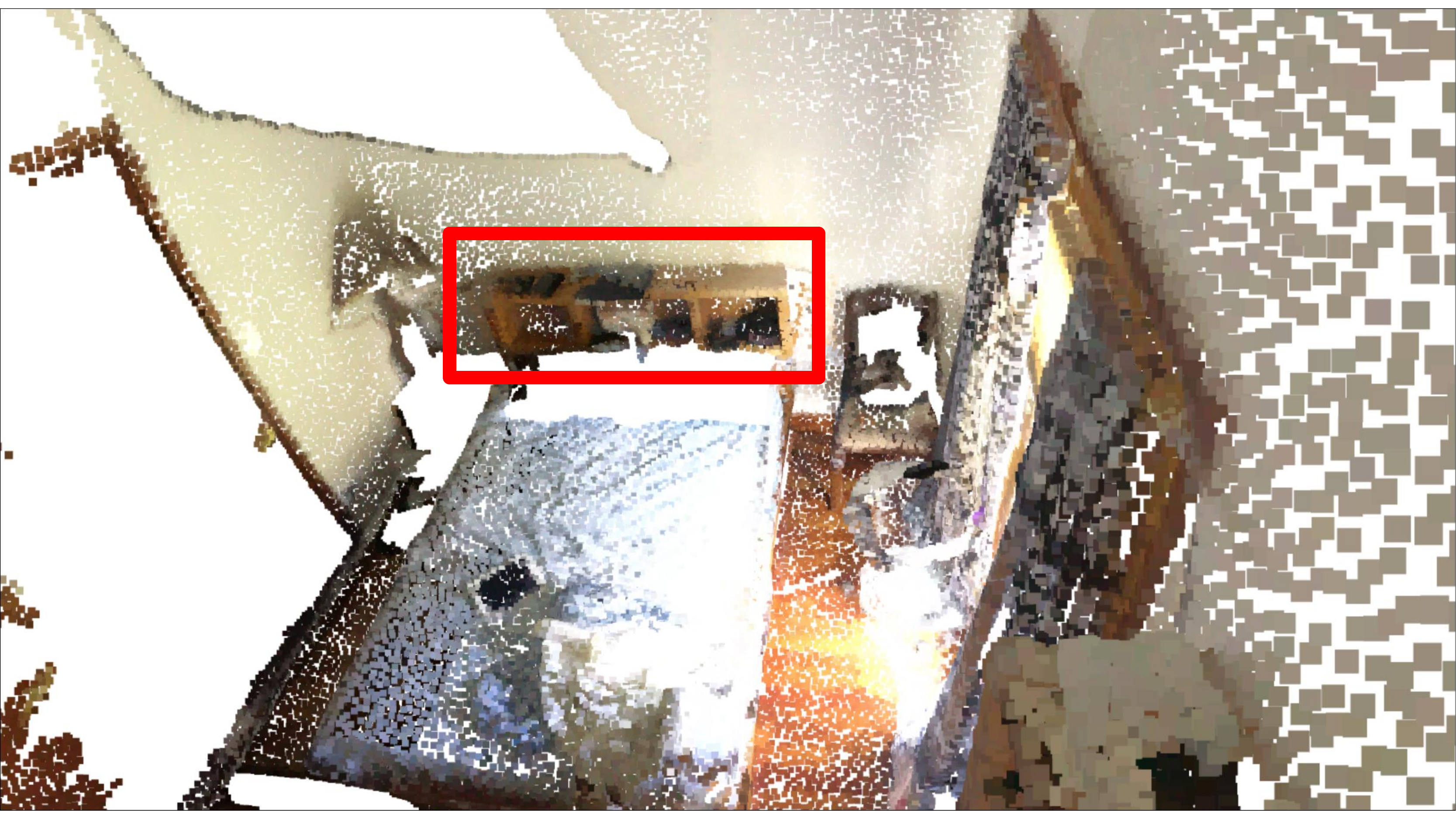} &
\includegraphics[width=0.190\textwidth,page=2]{img/scannetv2_our_vs_gt_vs_scn/203_oa0.7819_oa20.8374.pdf} &
\includegraphics[width=0.190\textwidth,page=3]{img/scannetv2_our_vs_gt_vs_scn/203_oa0.7819_oa20.8374.pdf} & 
\includegraphics[width=0.190\textwidth,page=4]{img/scannetv2_our_vs_gt_vs_scn/203_oa0.7819_oa20.8374.pdf} \vspace{-1mm} & \\
 Input Scenes & GT Labels & \cite{3DSemanticSegmentationWithSubmanifoldSparseConvNet} & \cite{3DSemanticSegmentationWithSubmanifoldSparseConvNet}+PNAL\\

\end{tabular}
}
\caption{From left to right: Scenes in ScanNetV2 validation set, GT labels given by ScanNetV2, predictions of \cite{3DSemanticSegmentationWithSubmanifoldSparseConvNet}, and predictions of \cite{3DSemanticSegmentationWithSubmanifoldSparseConvNet}+PNAL. Ours get more reasonable labels than GT labels.}
\label{Fig:scannetv2test}
\vspace{-0.5em}
\end{figure*}

\begin{figure*}[!ht] 
\setlength{\belowcaptionskip}{2pt} 
\centering
\setlength{\tabcolsep}{0.2mm}{
\begin{tabular}{cccccc}
\includegraphics[width=0.156\textwidth]{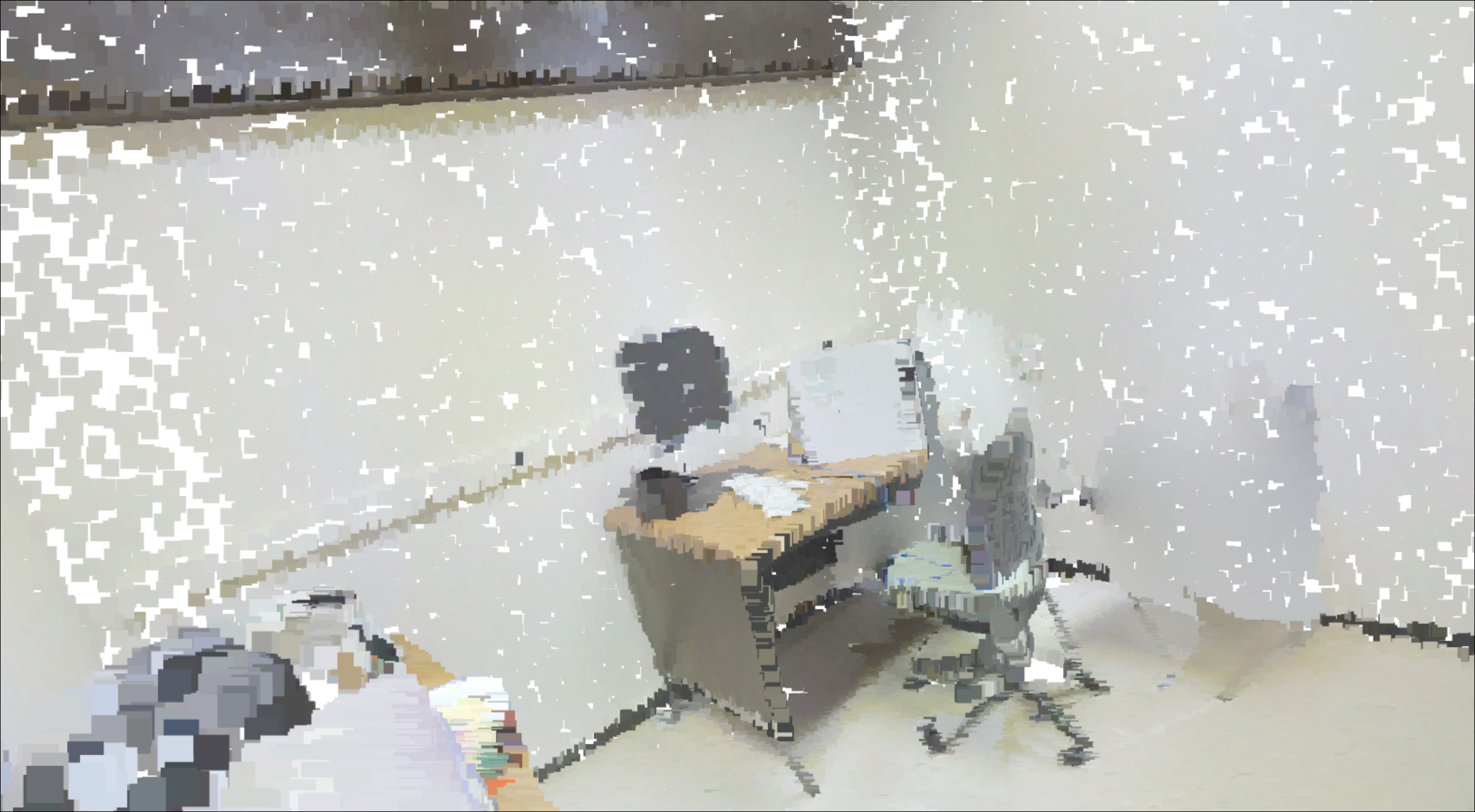} &
\includegraphics[width=0.156\textwidth]{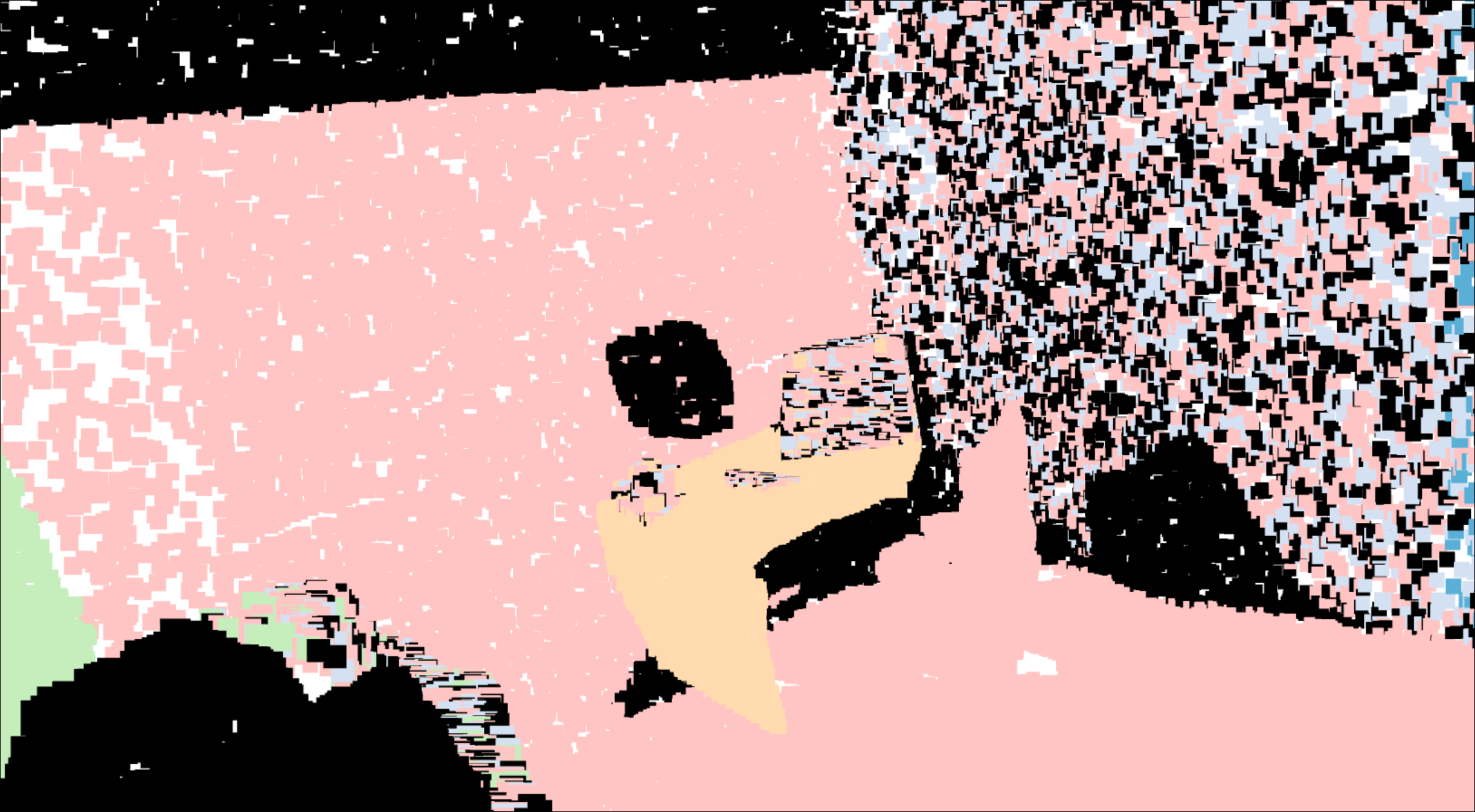} &
\includegraphics[width=0.156\textwidth]{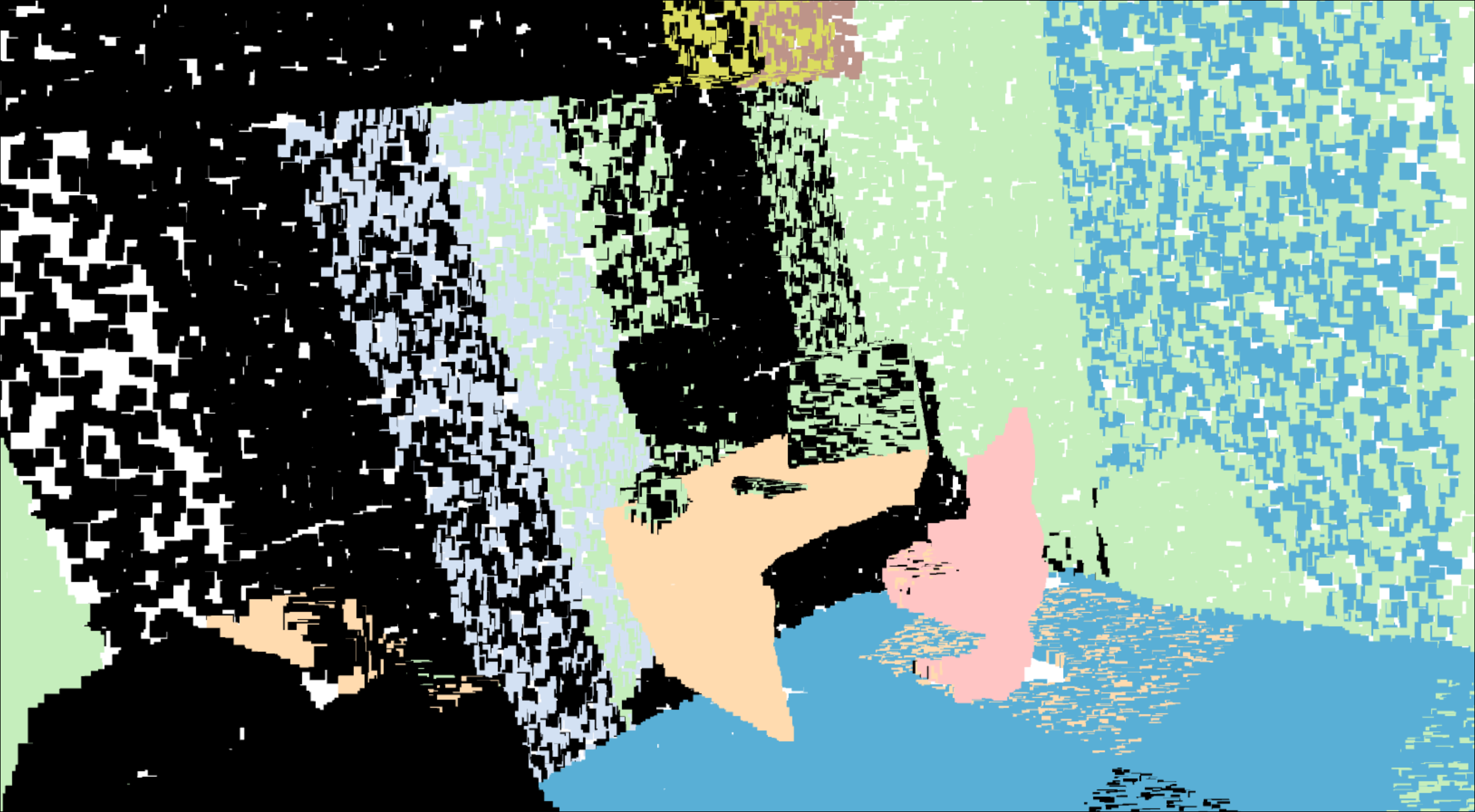} & 
\includegraphics[width=0.156\textwidth]{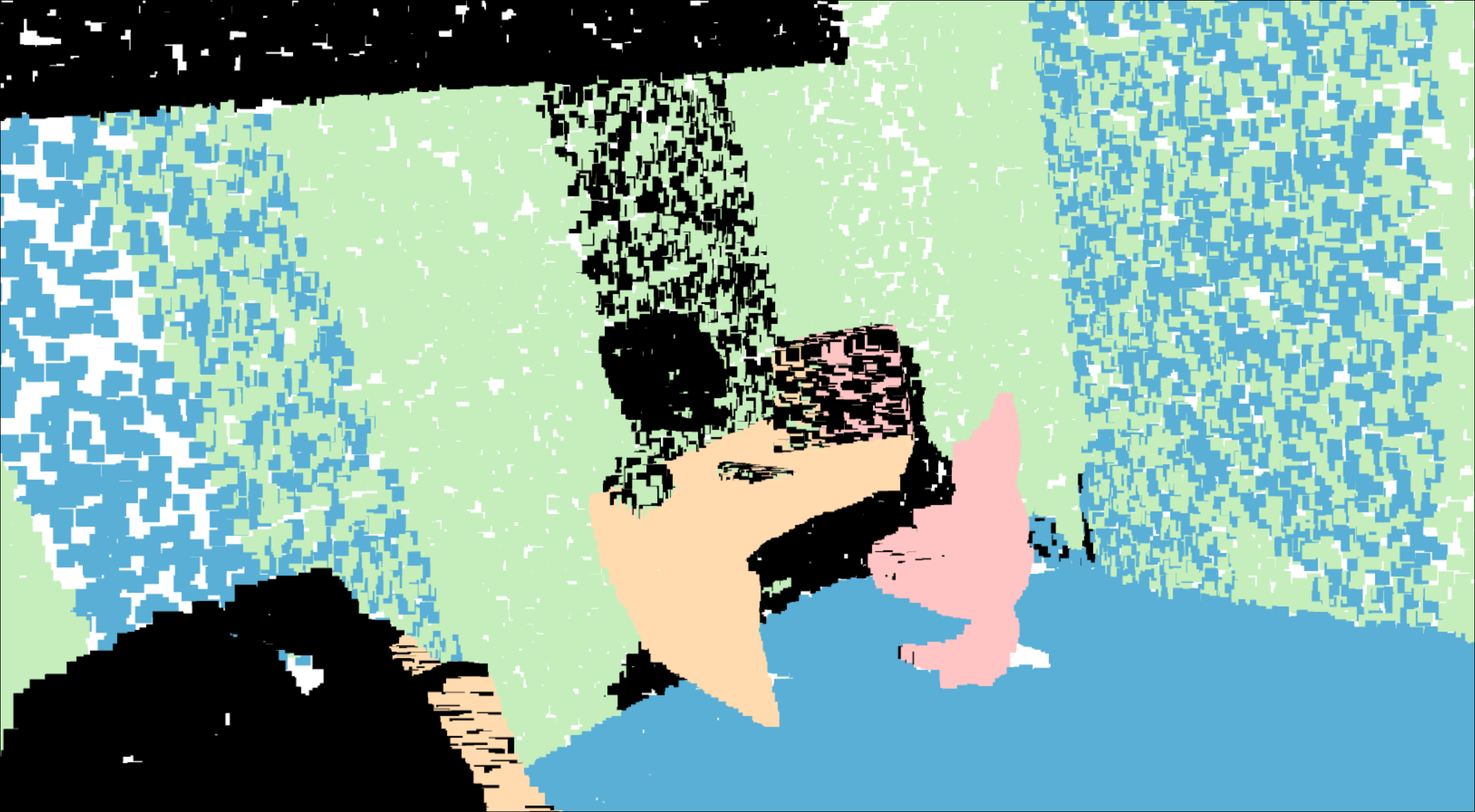} &
\includegraphics[width=0.156\textwidth]{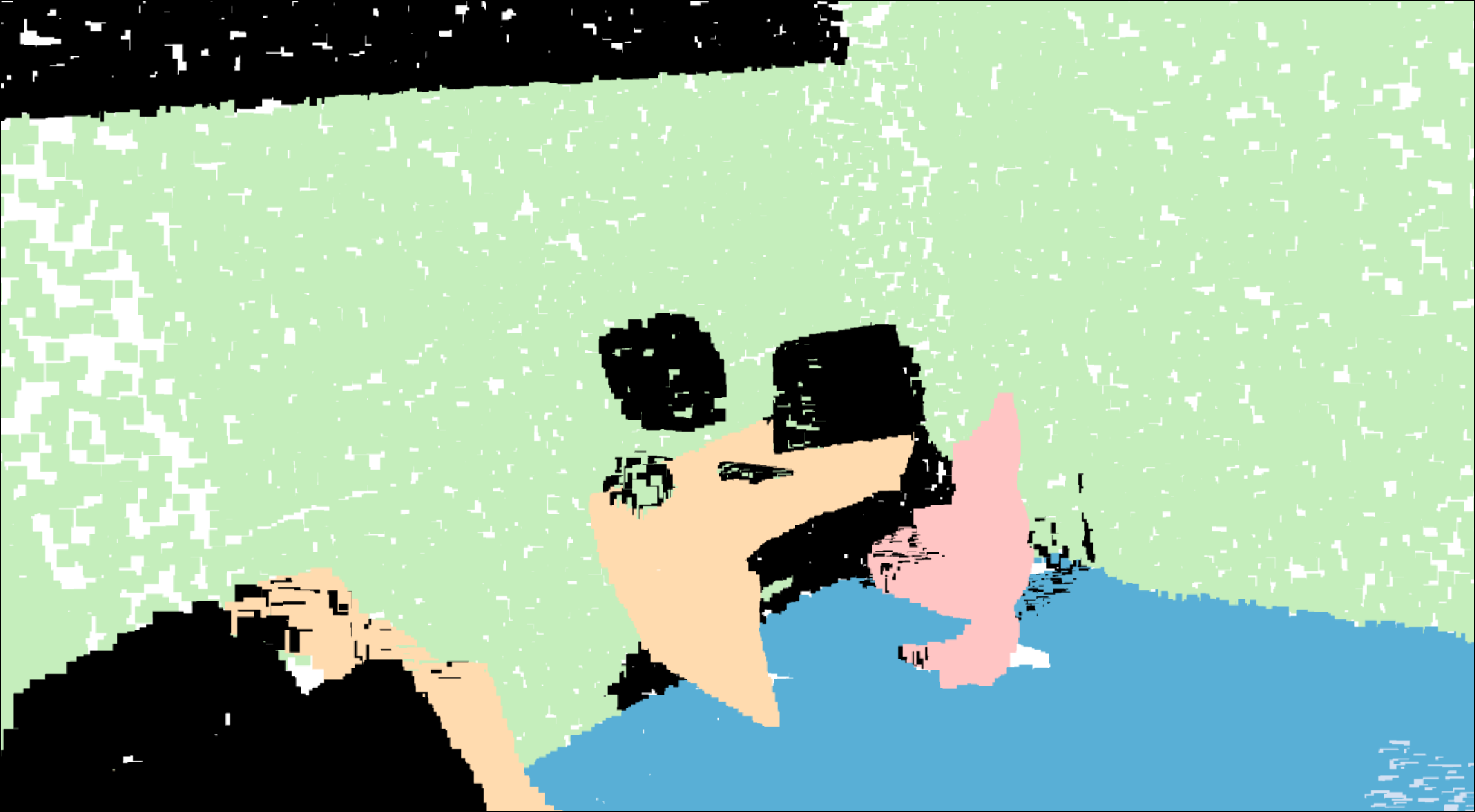} & 
\includegraphics[width=0.156\textwidth]{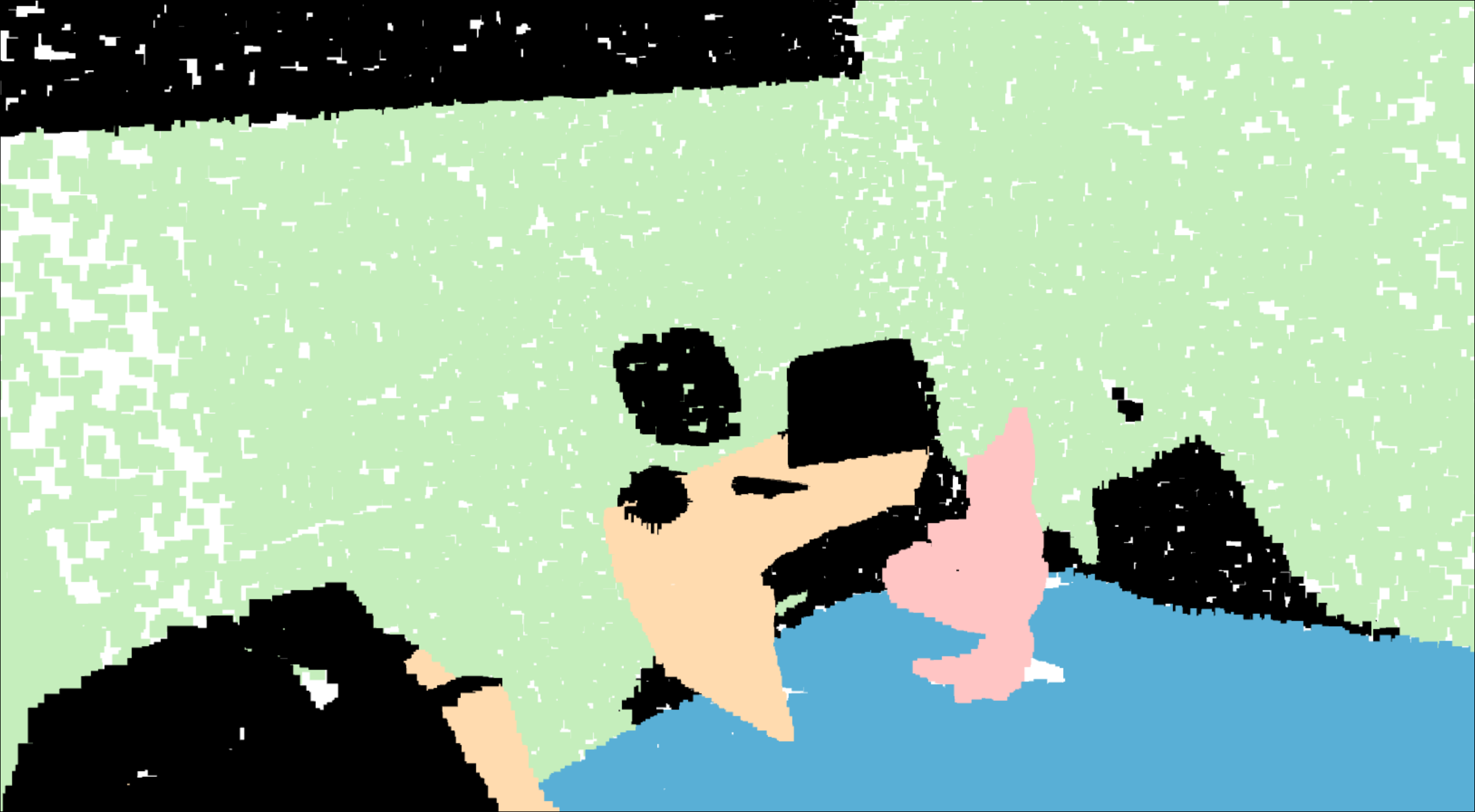} \vspace{-1mm} \\ 
 Input Scenes & Noisy Labels & 5 & 15 & 29 & GT Labels\\

\end{tabular}
}
\caption{Comparison of the corrected labels in epochs 5, 15, and 29 on the 60\% symmetric noise S3DIS. From left to right are the input point cloud with RGB, the corresponding noisy label, the label correction in epochs 5, 15, and 29, and the ground-truth label.}
\label{Fig:s3dis_corrprocess}
\vspace{-1em}
\end{figure*}

Table \ref{Tab:scannetv2} shows the performances on ScanNetV2 validation set with SparseConvNet \cite{3DSemanticSegmentationWithSubmanifoldSparseConvNet} as the backbone network.
Although the results of our method do not show a great advantage, we believe this is due to the fact that there are still mislabels in the validation set of ScanNetV2. 
Therefore, we show some examples from the validation set in the Fig. \ref{Fig:scannetv2test}, where ours gets more reasonable results than the baseline method and even than GT.

\begin{table}[]
\centering
\scalebox{0.875}{
\begin{tabular}{@{}c|cccc@{}}
\toprule
\multicolumn{1}{c|}{\multirow{2}{*}{Methods}} & \multicolumn{2}{l}{\begin{tabular}[c]{@{}c@{}}real-world noisy \\ ScanNetV2\end{tabular}} & \multicolumn{2}{l}{\begin{tabular}[c]{@{}c@{}}our re-labeled\\ ScanNetV2\end{tabular}} \\ \cmidrule(l){2-5} 
\multicolumn{1}{c|}{}                         & mIoU                                        & OA                                          & mIoU                                       & OA                                        \\ \cmidrule(r){1-5}
SparseConvNet\cite{3DSemanticSegmentationWithSubmanifoldSparseConvNet}                                     & 0.7250                                      & 0.8928                                      & 0.7103                                     & 0.8807                                    \\
SparseConvNet\cite{3DSemanticSegmentationWithSubmanifoldSparseConvNet}+PNAL                                & 0.7298                                      & 0.8979                                      & 0.7416                                     & 0.9211                                    \\ \bottomrule
\end{tabular}
}
\caption{The mIoU and OA comparison on real-world noisy ScanNetV2 validation set and our re-labeled ScanNetV2 validation set.}
\label{Tab:scannetv2}
\vspace{-1em}
\end{table}

For a more rigorous comparison, we further tested on the fully relabeled clean validation data mentioned before, and reported the results in Table \ref{Tab:scannetv2}. Our method achieves significant performance gain, which demonstrates the effectiveness of our method on the real-world noisy dataset. In contrast, the performance of the baseline SparseConvNet degrades, indicating its overfitting of the label noise, which affects the performance on fully clean data.

\subsection{Ablation Study}

\noindent\textbf{Component Ablation Study.}
\begin{table}[!t]
\centering
\scalebox{0.84}{
\begin{tabular}{@{}c|cccccc@{}}
\toprule
Metric & \begin{tabular}[c]{@{}c@{}}GT \\ Instance\end{tabular} & \begin{tabular}[c]{@{}c@{}}w/o \\ Voting\end{tabular} & $\gamma$=1 & $\gamma$=2 & $q$=8  & \begin{tabular}[c]{@{}c@{}}DGCNN\\ +PNAL\end{tabular} \\ \midrule
OA      & \textbf{0.8287}                                        & 0.8011                                                & 0.8110     & 0.8209     & 0.7704 & 0.8236                                                \\ \bottomrule
\end{tabular}
}
\caption{OA Comparison on $60\%$ symmetric noise S3DIS dataset.} 
\label{Tab:ablation}
\end{table}
All the results in Table. \ref{Tab:ablation} are on $60\%$ symmetric noise.
The first column reports the results of PNAL with GT instance instead of clustering for label correction, which represents the upper limit of our results. Compared with it, the cluster based results in the last column have only a small drop, which can illustrate the feasibility of using cluster as an alternative to GT instance label.
In the second column, we omit the cluster-level voting step and perform label correction point-wisely, without considering the label consistency between nearby points. The results show a $2.25\%$ decrease over our full method, demonstrating the effectiveness of our proposed cluster-level label correction.
In the third and fourth columns, we try different values of $\gamma$, where $\gamma=1$ is the most greedy case, i.e., the winner label is the top reliable label. We observe no significant performance drop with different $\gamma$ values, implying our method is not sensitive to the choice of $\gamma$. We use $\gamma =4$ in our setting. 
In the fifth column we adjust the history length $q$ to 8, and note that $E_{warm-up}$ is also increased to 8, due to the constraint of $q$. We can observe a significant decrease in performance. More analysis is given in the next paragraph.

\noindent\textbf{Robustness to $E_{warm-up}$.}\label{subsec:abl1_Ewarmupvsnoiserate}
Table \ref{Tab:Ewarmupvsnoiserate} reports the results of our method at different $E_{warm-up}$ under symmetric noise of different noise rates ($20\%, 40\%, 60\%$). Our performances are optimal and robust, with $E_{warm-up}=5$ as we recommend, for all noise rates, demonstrating that $E_{warm-up}$ is not sensitive to noise rate variations.
The larger the noise rate, the larger the performance drop can be observed if $E_{warm-up}$ increases. We can conclude that the larger the $E_{warm-up}$, the more noisy data the network fits, which can make noise-cleaning difficult on data with large noise rates.
Comparing the result in the second row with the fifth column in Tab.~\ref{Tab:ablation}, we found that although increasing the history length brings a performance decrease, this mainly comes from the effect of the increased $E_{warm-up}$.
\begin{table}[t]
\centering
\setlength{\tabcolsep}{5.0mm}
{
\begin{tabular}{@{}c|ccc@{}}
\toprule
Noise Rate $\tau$ & 20\%   & 40\%   & 60\%   \\ \midrule
$E_{warm-up}=5$         & \textbf{0.8569} & \textbf{0.8378} & \textbf{0.8236} \\
$E_{warm-up}=8$         & 0.8422 & 0.8247 & 0.7851 \\
$E_{warm-up}=11$        & 0.8343 & 0.8009 & 0.7812 \\ \bottomrule
\end{tabular} 
}
\vspace{-1.2mm}
\caption{The OA of PNAL at different $E_{warm-up}$ and noise rates.} \label{Tab:Ewarmupvsnoiserate}
\end{table}

\noindent\textbf{Robustness to Clustering Methods and Granularity.}\label{subsec:abl2_vsEpsandGMM}
In Table~\ref{tab:granularity}, the first three columns report the results of PNAL under different clustering granularities ($\varepsilon$ is $0.015, 0.018, 0.021$), and the last two columns report the results of ours under other type of clustering (GMM, spectral). They show close results, which demonstrates that our method is robust to a certain range of cluster granularities and is not sensitive to the clustering method used.

\begin{table}[t]
\centering
\setlength{\tabcolsep}{0.5mm}{
\begin{tabular}{@{}c|ccccc@{}}
\toprule
\begin{tabular}[c]{@{}c@{}}Clustering \\ Methods\end{tabular} & \begin{tabular}[c]{@{}c@{}}DBSCAN \\ $\varepsilon$=0.015\end{tabular} & \begin{tabular}[c]{@{}c@{}}DBSCAN \\ $\varepsilon$=0.018\end{tabular} & \begin{tabular}[c]{@{}c@{}}DBSCAN\\ $\varepsilon$=0.021\end{tabular} & GMM & spectral    \\ \midrule
OA                                                            & 0.8206                                                      & \textbf{0.8236}                                             & 0.8159                                                     & 0.8178 & 0.8162 \\ \bottomrule
\end{tabular}
}
\vspace{-1.2mm}
\caption{The OA of our method at different granularity of clusters and different clustering methods.}
\label{tab:granularity}
\vspace{-1.5em}
\end{table}

\vspace{-1.5em}
\subsection{Correction Process Analysis}\label{subsec:corrprocess}
\begin{figure}[t]
\vspace{-0.6mm}
\includegraphics[width=0.45\textwidth]{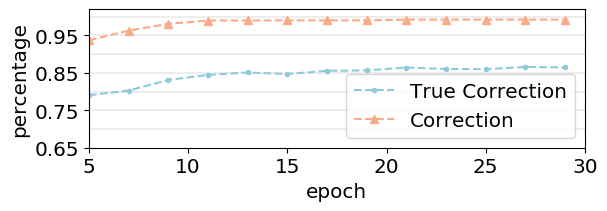}
\vspace{-2.6mm}
\caption{The percentages of points
with replaced label (denoted as Correction) and the percentage of correctly corrected points (denoted as True Correction).}
\label{Fig:corrector_percent}
\vspace{-1.5em}
\end{figure}

\vspace{-0.5em}
Fig. \ref{Fig:s3dis_corrprocess} shows the visualization of the label correction process by the PNAL during training. We can find that as the training goes, the overall labeling errors in the training set tends to decrease and gradually approaches the clean ground-truth label. Typically, label errors with large areas (e.g., floors, walls, ceilings) are corrected first. As training proceeds, PNAL gradually explores the entire dataset and tries to correct difficult and small objects. As given in Fig. \ref{Fig:corrector_percent}, the percentage of points with replaced label increases from $0.936$ to $0.992$, while the percentage of correctly corrected points is close to $0.8$ from the beginning of the noise-cleaning stage and then gradually increases to $0.865$. It also shows that the PNAL correction process spreads to whole training set as the training proceeds. To note that we take the case of correcting to original label into account.

\section{Conclusion}
In this study, we propose PNAL, a new point cloud segmentation framework, to cope with the novel point cloud labeling noise problem. Unlike existing methods that focus on image classification, PNAL is noise-rate blind, in order to cope with the unique noise-rate variation problem in point cloud. We propose point-wise confidence selection, cluster-wise label correction and voting strategies to generate the best possible labels considering the correlation labels in local similar points. In addition, we re-labeled the validation set of a popular but noisy real-world scene point cloud dataset to make it clean, for rigorous experiment and for future research. Experiments demonstrate the effectiveness and robustness of our method on real-world noisy data and artificially created noisy public data.\\

\vspace{-1.0em}
\noindent\textbf{Acknowledgements:} We would like to thank Jiaying Lin for his generous help. This work was supported by the National Natural Science Foundation of China under Grant U20B2047 and the Shenzhen Basic Research General Program under Grant JCYJ20190814112007258.

{\small
\bibliographystyle{ieee_fullname}
\bibliography{egbib}
}

\end{document}